\documentclass[fleqn,10pt]{wlscirep}
\usepackage[utf8]{inputenc}
\usepackage[T1]{fontenc}
\usepackage{physics}
\usepackage{enumerate}
\usepackage{enumitem}
\usepackage{mathtools}
\newcommand{\imag}{\mathrm{i}}
\usepackage{hyperref}
\usepackage{enumitem}

\usepackage{placeins}

\title{Boosting Classification with Quantum-Inspired Augmentations}

\author[1]{Matthias Tsch\"ope}
\author[1,3]{Vitor Fortes Rey}
\author[2]{Sogo Pierre Sanon}
\author[1,3]{Nikolaos Palaiodimopoulos}
\author[1,3]{Paul Lukowicz}
\author[1,3,4,*]{Maximilian Kiefer-Emmanouilidis}

\affil[1]{German Research Center for Artificial Intelligence (DFKI), Embedded Intelligence, Kaiserslautern, 67663, Germany}
\affil[2]{German Research Center for Artificial Intelligence (DFKI), Intelligent Networks, Kaiserslautern, 67663, Germany}
\affil[3]{RPTU Kaiserslautern-Landau, Department of Computer Science and Research Initiative
QC-AI, Kaiserslautern, 67663, Germany}
\affil[4]{RPTU Kaiserslautern-Landau, Department of Physics,
Kaiserslautern, 67663, Germany}
\affil[*]{maximilian.kiefer@rptu.de}


\begin{abstract}
Understanding the impact of small quantum gate perturbations, which are common in quantum digital devices but absent in classical computers, is crucial for identifying potential advantages in quantum machine learning. While these perturbations are typically seen as detrimental to quantum computation, they can actually enhance performance by serving as a natural source of data augmentation. Additionally, they can often be efficiently simulated on classical hardware, enabling quantum-inspired approaches to improve classical machine learning methods.

In this paper, we investigate random Bloch sphere rotations, which are fundamental SU(2) transformations, as a simple yet effective quantum-inspired data augmentation technique. Unlike conventional augmentations such as flipping, rotating, or cropping, quantum transformations lack intuitive spatial interpretations, making their application to tasks like image classification less straightforward. While common quantum augmentation methods rely on applying quantum models or trainable quanvolutional layers to classical datasets, we focus on the direct application of small-angle Bloch rotations and their effect on classical data. We argue that these transformations can be implemented efficiently on classical hardware, offering more precise analysis of augmentation effects than is currently possible on noisy quantum devices. Using the large-scale ImageNet dataset, we demonstrate that our quantum-inspired augmentation method improves image classification performance, increasing Top-1 accuracy by 3\%, Top-5 accuracy by 2.5\%, and the F$_1$ score from 8\% to 12\% compared to standard classical augmentation methods. Finally, we examine the use of stronger unitary augmentations. Although these transformations preserve information in principle, they result in visually unrecognizable images with potential applications for privacy computations. However, we show that our augmentation approach and simple SU(2) transformations do not enhance differential privacy and discuss the implications of this limitation.
\end{abstract}

\begin{document}

\flushbottom
\maketitle

\thispagestyle{empty}

\section*{Introduction}\label{sec1}

In classical Machine Learning (ML), the performance of the algorithms strongly depends on the availability of high-quality data \cite{Khalifa2022}. A lack of sufficient or bad training data can lower the model performance, reduce accuracy and disturb convergence \cite{Khalifa2022,bacstanlar2014}. The introduction of data augmentation techniques \cite{van2001art} improved the performance and generalization ability of classical ML models without the need for more high-quality annotated data \cite{shorten2019survey,xu2023comprehensive,mikolajczyk2018data}. Quantum Machine Learning (QML) similarly faces comparable data limitations \cite{biamonte2017quantum, Perrier2022}. Recent studies have proposed utilizing quantum circuits for data augmentations \cite{Chalumuri2022, Apak2024, GAO2022109515, baglio2024dataaugmentationexperimentsstylebased} to enhance the training of Quantum Neural Networks (QNNs) \cite{Chalumuri2022, Apak2024, Don2025} as well as hybrid and classical Convolutional Neural Networks (CNNs) \cite{Henderson2019, Mari2020, Chalumuri2022,gordienko2025hnn}. However, certain augmentations are directly problem-specific--for instance, in image signal transfer tasks \cite{palaiodimopoulos2024quantum}-- or may even hinder the expressivity of QNNs when they correspond to effectively random weights in a trained network \cite{werner2024disqu}. 

Despite the growing interest in data augmentation via quantum augmentation algorithms, the benefits of simple quantum-inspired transformations remain largely unexplored \cite{palaiodimopoulos2024quantum}. Their potential applicability to classical machine learning is also not yet well understood.  In parallel, current quantum hardware is constrained by a limited number of qubits and high levels of noise, which hinders the systematic analysis and practical deployment of QML algorithms \cite{Preskill2018}. One promising strategy to address these limitations is the development of quantum-inspired algorithms--classically implementable methods grounded in quantum principles \cite{Chia2018, Arrazola2020, srinivasan2024qseg, srinivasan2025comparative}.

These algorithms can either leverage quantum concepts to improve the performance or provide insight into classical problems from a quantum perspective \cite{Papalitsas2021}, or repurpose tools used to efficiently solve quantum systems \cite{huynh2023quantum}. Within this framework, data augmentation remains a key application area, particularly, in advanced computer vision models based on contrastive learning, where composite transformations play a central role. In the classical setting, SimCLR \cite{chen2020simple} demonstrated the importance of such compositions for effective feature learning. Quantum and quantum-inspired extensions have since emerged, such as Q‑SupCon \cite{Don2025}, which extends supervised contrastive learning with a dedicated quantum data‑augmentation \cite{Chalumuri2022} circuit. Other works have used trainable parametrized quantum circuits as layers within hybrid models combined with classical data augmentation \cite{chen2025quantummultimodalcontrastivelearning, Yanhui2024, REN2025100574, Jaderberg_2022}, or have focused specifically on enhancing the projection head \cite{kankeu2025quantuminspiredembeddingsprojectionsimilarity, Yu2024}. However, most reported quantum advantages in data augmentation are demonstrated on simple benchmark datasets like MNIST \cite{bowles2024better}, and its unclear if the same procedures adapt to significantly harder datasets.

In this work, we seek to address this gap by demonstrating that random Bloch rotations—a type of transformation that commonly occurs as a side effect of noise in real quantum hardware—can be leveraged systematically as a data augmentation technique for classical image classification tasks. To assess the effectiveness of this approach, we evaluate both classical and quantum-inspired augmentations, as well as their sequential combination. In our setup, classical augmentations are applied first, followed by quantum-inspired transformations. We assess performance on the ImageNet dataset using a ResNet-34 architecture. ResNet, introduced by Kaiming He et al. in 2016 \cite{he2016deep}, marked a significant breakthrough in deep learning by addressing the vanishing gradient problem through residual connections. Instead of learning direct input–output mappings, residual blocks learn functions relative to the input, enabling more effective training of deeper networks by facilitating gradient flow during backpropagation. Our results indicate that the proposed augmentation methods can improve classification performance improving Top-1 accuracy by 2\%, Top-5 accuracy by 1\%, and the F$_1$ score remarkably from 8\% to 12\% compared to standard and only classical augmentation methods. 

Although the resulting quantum-inspired algorithm is motivated by quantum gate operations, it can be fully understood within a classical framework and executed efficiently on a classical computer with a computational complexity of $\mathcal{O}(N\mathrm{log}(N))$, where $N$ is the dimension of the flattened image. Additionally, the presented algorithm may be implemented in optical setups of waveguides \cite{Macho-Ortiz2021, palaiodimopoulos2024quantum} which in principle can be beneficial for energy-efficient computing in embedded systems. The presented augmentation algorithm can work with classical data as well as with quantum data, and in the latter one does preserve all entanglement properties of underlying quantum data.

The manuscript is structured as follows: In section \hyperref[sec:qca]{"Quantum and Classical Augmentations"}, we first introduce the quantum and classical augmentation techniques and discuss their effects and associated invariants. We then propose augmentation methods that integrate both classical and quantum approaches. In section \hyperref[sec:m]{"Methods"}, we describe the machine learning model and the dataset, and provide detailed information about the training pipeline. In section \hyperref[sec:results]{"Results"} we present and discuss our results. Our conclusions are summarized in section \hyperref[sec:discuss]{"Discussion"}. Additional considerations and results are deferred to the \hyperref[sec:sup]{"Supplementary Material"}.

\section*{Quantum and Classical Augmentations} \label{sec:qca}
Before we can present the augmentation routine we first discuss how we represent the data as a quantum-inspired state. For this task the optimal quantum embedding is common amplitude embedding \cite{1mottonen2005}, which up to the normalization factor, is the closest representation for real or complex classical data, with dramatic reduced memory effort in the quantum case with $n=\log_2 N$ necessary qubits to encode a flattened dataset of size $N$. In current noisy devices \cite{preskill2018quantum, bowles2024better} the overhead to write and read the data \cite{1mottonen2005} significantly hinders the widespread application of amplitude encoding, although recently much effort has been made to reduce these efforts significantly for classical data \cite{araujo2023low}. Given any two-dimensional dataset (in our case an image), we can express it as
\begin{equation}
    \mathrm{Image}= \sum_{i,j}^{W,L} C_{ij} \ket{i}\bra{j}, \qquad \ket{\mathrm{Image}}_{f} = \sum_{i,j}^{W,L} C_{ij} \ket{i} \otimes \ket{j},
\end{equation}
where $\ket{i}, \ket{j}$ are basis vectors (computational basis) forming the outer product $\ket{i}\bra{j}$ and $C_{ij}$ are the coefficients of the corresponding two-dimensional array indexed by $i,j$. Then $\mathrm{\ket{Image}}_{f}$ corresponds to the flattened image and represents the non-normalized amplitude encoded quantum state where '$\otimes$' is the Kronecker product. The corresponding quantum state would then need to be normalized by the Euclidean norm $N=\sqrt{\sum_{i,j}^{W,L} C_{ij}C_{ij}^*}$. We avoid this step since our algorithm is quantum-inspired and we later apply a min-max normalization which is better suited for our classification task. From a quantum perspective, the image has been encoded as a probability distribution $p(x)$, where $x$ represents the register address and the actual pixel value corresponds to the probability how often we measure the address state in our circuit.

Alternative schemes\cite{Lisnichenko2022}, such as angle encoding, are not suitable for our approach, as they require one qubit per feature, making them inefficient for high-dimensional data on quantum and classical hardware. Moreover, they do not preserve the relative magnitudes or the global structure of the data vector. Similarly, the Flexible Representation of Quantum Images (FRQI), which is specifically designed for spatially structured image data, and introduces a pixel-based representation that is not meaningful to preserve in our context. Furthermore, the here presented augmentation in the form of Bloch rotations would resemble rather a classical augmentation, similar to additive uniform noise.

\subsubsection*{Quantum Augmentation}   
The augmentation makes heavy use of fundamental Bloch rotations which are given by 
\begin{align}
\begin{split}
    R_X(\theta)&= \exp(-\mathrm{i}\theta  \hat X/2) = \left(\begin{array}{cc}
        \cos{\theta/2} & -\imag\sin{\theta/2} \\
        -\imag\sin{\theta/2} & \cos{\theta/2}
    \end{array}\right), \quad 
    R_Y(\theta)= \exp(-\mathrm{i}\theta  \hat Y/2) = \left(\begin{array}{cc}
        \cos{\theta/2} & -\sin{\theta/2} \\
        \sin{\theta/2} & \cos{\theta/2}
    \end{array}\right), \\
    R_Z(\theta)&= \exp(-\mathrm{i}\theta \hat  Z/2) = \left(\begin{array}{cc}
        \exp(-\imag \theta /2)& 0 \\
        0 & \exp(\imag \theta /2)
    \end{array}\right) ,
\end{split}
\end{align}
where $\hat X,\hat Y, \hat Z$ are the Pauli operators. 

\begin{figure}[ht]
    \centering
    \includegraphics[scale=0.7]{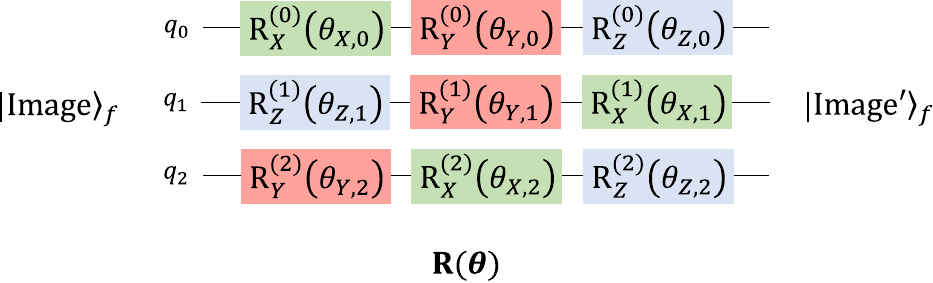}
    \caption{An example of the proposed augmentation circuit for three qubits.}
    \label{Circuit}
\end{figure}

For each qubit $i$ and axis $j \in (X, Y, Z)$, a rotation angle $\theta_{i,j}$ is independently sampled from the uniform distribution $[0, \Theta]$, where $\Theta$ represents the strength of the rotation. In addition, the $j$'s are randomly drawn from the ordered tuple $(X, Y, Z)$ determining the order in which single-qubit rotations around the $X$, $Y$, and $Z$ axes are applied. The quantum part of the augmentation is then given by
\begin{equation} \label{eq:qrot}
    \ket{\mathrm{Image}'}_{f} = QR_{XYZ}(\ket{\mathrm{Image}}_{f})= \mathbf{R}(\theta) \ket{\mathrm{Image}}_{f}=\bigotimes_{i=0}^{n-1} \sideset{}{'}\prod_{j \in (X,Y,Z) } R_j^{(i)}(\theta_{j,i}) \ket{\mathrm{Image}}_{f},
\end{equation}
where the $\sideset{}{'}\prod$ indicates the non-commutative product of rotation gates drawn from the random ordered tuple j. An example of a three-qubit circuit implementing this augmentation is shown in Fig.~\ref{Circuit}. We also introduce the subsets $QR_X, QR_Y, QR_Z$, which reduce Eq. (\ref{eq:qrot}) to a single direction quantum rotation. As part of the augmentation we will evaluate the effects of an element-wise absolute (abs()) or real (real()) values of the state $\ket{\mathrm{Image}'}_{f}$. Taking the absolute value corresponds to accessing measurement outcomes or, more precisely, the square roots of output probabilities. Obtaining the real part of a quantum state can be achieved through a Hadamard test \cite{kubo2021variational} on a quantum device. However, it is important to note that these operations are highly inefficient on current quantum computers and highlight that, in scenarios where augmentation is the sole objective—rather than a component of a full quantum algorithm, the evaluation on a classical computer is likely to be more practical and effective.

The randomness in the direction of the rotations $(X,Y,Z)$ is to minimize the effect of random values close to each other rotated to the same point. Thus, we explore all 6 possible Trait-Bryan angles with three different extrinsic rotation axes (special case of Euler angles with three different axes). However, two remarks should be made: First, although we apply multiple rotations to different qubits the corresponding group for the rotation stays SU(2) as no entangling gates are used in the augmentation procedure \cite{de2018simple} and thus no rotation in SU(N). And, furthermore, the classical interpretation that we rotate an actual image via a certain rotation axes is not the case. We should rather think of a mixer layer most common in quantum approximate optimization algorithms (QAOA)  where Bloch rotations redistribute amplitudes among feasible states, enabling the algorithm to escape local minima and explore the solution space \cite{moll2018quantum, zhou2020quantum}. 

\subsubsection*{Algorithm Description and Complexity }

The efficient application of the algorithm is due to its SU(2) group, enabling us to apply the quantum rotation gates onto the qubits separately instead of creating a large $NxN$ matrix. Thus, a fully separable state stays fully separable after the transformation\cite{Bley2024}. Without loss of generality, we assume $N=2^n$, where n is the number of qubits, if this is not the case we can use padding with zeros to fulfil this condition. The following steps need to be repeated $n$ times and $i=0$:

\begin{itemize}
    \item Reshape state $\ket{\mathrm{Image}}_{f}$ into a matrix $\mathrm{Image}_{f}$ of size $2\times N/2$.
    \item Apply a matrix multiplication of the Bloch rotations on the corresponding matrix $\mathrm{Image}_{f} \leftarrow \sideset{}{'}\prod_{j \in (X,Y,Z) } R_j(\theta_{j,i}) \mathrm{Image}_{f}$.
    \item Transpose $\mathrm{Image}_{f}$ and flatten the image back to the state $\ket{\mathrm{Image}}_{f}$ and $i+1$.
\end{itemize}
Afterwards, we either keep the full state or evaluate the element-wise $abs()$ or $real()$ values.  

The complexity of this operation is simply given by $\mathcal{O}(N\log(N))$ as the augmentation needs to be applied $n=\log(N)$ times and the corresponding matrix multiplication of $2\times 2$ Bloch rotations with the reshaped image dimension $2\times N/2$ results in a complexity of $\mathcal{O}(N)$. Note, that the application of $QR_Z$ can be done even more efficient, as the operation is given by a diagonal matrix and the complexity reduces to element-wise multiplication of two $N$ sized vectors $\mathcal{O}(N)$.

\subsection*{Effects of the Augmentations and Invariants}
To evaluate the effects of an abstract data augmentation, it is crucial to understand what changes we would expect and what classical analogue we can find which shows a similar effect. The obvious invariant of the unitary transformation is the Euclidean norm. However, this global invariant does not correspond to any rich property of an image. Thus, we focus on a more suitable invariant for our interpretations: the singular values of an image.
We begin by showing that the singular values of an image correspond to invariants under the quantum-inspired transformation. Specifically, we start with the Schmidt decomposition (equivalently, the singular value decomposition) of the embedded state $\ket{x}$. Without loss of generality, we partition the state into two equally sized Hilbert spaces, each corresponding to $n/2$ qubits, such that
\begin{equation}
    \ket{x}= \sum_k c_k\ket{\xi_k}\ket{\theta_k}, 
\end{equation}
where $c_k$ are the real and semi-positive singular values of the image, and $\ket{\xi_k}, \ket{\theta_k}$ are the corresponding orthonormal basis states of the Hilbert spaces associated with the first $n/2$ and last $n/2$ qubits. Applying Eq.~(\ref{eq:qrot}) we obtain
\begin{align}
    QR_{XYZ}(x)=\sum_k c_k\bigotimes_{i=0}^{n-1} \sideset{}{'}\prod_{j \in (X,Y,Z) } R_j^{(i)}(\theta_{j,i}) \ket{\xi_k} \bigotimes_{i=n}^{2n-1} \sideset{}{'}\prod_{j \in (X,Y,Z) } R_j^{(i)}(\theta_{j,i}) \ket{\theta_k}=\sum_j c_k\ket{\xi_k'}\ket{\theta_k'},
\end{align}
which shows that $c_k$ remained unchanged, implying that operations which do not act between Hilbert spaces cannot change the singular values of the corresponding state. In this sense, the analogy reflects the principle that single-qubit operations cannot alter entanglement.  
Having established that the singular values remain invariant under the quantum transformation, we now examine how small perturbations might affect this invariance, by taking the aforementioned element-wise absolute value corresponding to measurement outcomes or, by evaluating the real part of the amplitudes.  Both of these operations, as they involve measurement, inherently disturb the quantum state—potentially destroying entanglement and, consequently, altering the singular values of the corresponding image. Since the augmentation does not create any new "entanglement" it is the inherent "entanglement" from the initial amplitude encoded image $\ket{\mathrm{Image}}_\mathrm{f}$ which is altered, and therefore the singular values of $\ket{\mathrm{Image}}_\mathrm{f}$ which are disturbed. Furthermore, without any quantum rotation $abs()$ and $real()$ would not change any image property.
 
Although singular values and their distributions are generally not expressive enough to fully capture the effects of transformations --since infinitely many images can share the same singular values while differing in their singular vectors-- the effects considered here can be evaluated when perturbations (the rotation angles) are small. As a result, these changes do not significantly alter the visual content of the image. However, changes in the singular values may reveal which aspects of the image this abstract augmentation might highlight or disguise. To explore this, we selected two sets of classical augmentations: one that, by design, preserves the singular values of the image, and another that deliberately perturbs them.

\subsection*{Classical Augmentations}
The set of augmentations that preserve the singular value spectrum, includes horizontal flipping and perfect rotations. The flip operation, denoted $F(x)$, corresponds to a horizontal reflection in which the image is mirrored from left to right. While the perfect rotation, denoted $PR(x)$, utilizes a standard rotation matrix to rotate the image about the origin. The transformation for arbitrary coordinates $u$ and $v$ is given by
\begin{equation} \label{eq:cr}
\begin{pmatrix}
u' \\
v'
\end{pmatrix}
=
\begin{pmatrix}
\cos\theta & -\sin\theta \\
\sin\theta & \cos\theta
\end{pmatrix}
\begin{pmatrix}
u \\
v
\end{pmatrix}.
\end{equation}
In the case of perfect rotations, the angle $\theta$ is restricted to multiples of $\pi/2$. When the angle is arbitrary, the transformation is referred to as a classical rotation and is denoted by $CR(x)$. It's also important to note that the classical rotation matrix corresponds to the Bloch rotation $R_Y(\theta)$. In contrast to the quantum case, where operations are performed on a quantum state representation of the image, the transformation here is applied directly to the pixel coordinates. 

The set of classical augmentations that do not preserve the singular value spectrum are small continuous rotations $CR(x)$, Gaussian noise $GN(x)$, and cropping $C(x)$. Gaussian noise is applied by sampling random values from a normal distribution with mean $\mu = 0$ and standard deviation $\sigma = 1$, and adding these values pixel-wise to the training images. This operation is then expected to reduce the importance of high-valued singular values and slightly emphasize the tail of the singular value spectrum. For cropping, the image is first enlarged, and a fixed-size region is then extracted from its center.
Here it is expected that, cropping on average will highlight the high-valued and the middle of the spectrum of the singular values and reduce the importance of the tail significantly.

What we aim to achieve is to evaluate both classical and quantum-inspired operations, individually, comparatively and in combination. To this end, in the next section we present our augmentation methods.


\subsection*{Augmentation Methods}\label{sec:am}
For the presented augmentation methods we will follow two limiting cases. The first is the weak Bloch rotations, where $\Theta$ is in the order of $10^{-2}$. Here, we can analyse the effects of the augmentation by comparing it to changes of the singular value spectrum. In the following, we will only present the weak case in the main text. Our findings indicate that stronger Bloch rotations do not enhance classification performance, even in the cases where one might expect an advantage due to information perseverance in the singular values, such as in differential privacy. We refer the reader to the \hyperref[sec:sup]{"Supplementary A"} for the description of the methods, proofs and results of the strong Bloch rotations.
\noindent Below, we provide a list of the augmentation methods employed for the weak rotating case, which are composed of the previously described operations: \par
\smallskip


\begin{description}[leftmargin=4cm, style=multiline]

\item[$BL$] Baseline (without any augmentation).

\item[$GN(x)$] Apply random Gaussian noise to the samples $x$.

\item[$F(PR(x))$] First apply a random perfect rotation (multiple of $90^\circ$), then randomly flip the image horizontally.

\item[$C(F(CR(x)))$] First, apply a random classical rotation by an angle between $-35^\circ$ and $+35^\circ$, then randomly flip the image horizontally, and finally crop the image to a fixed size.

\item[$real(QR_Y(x))$] Apply quantum rotation on the Y-axis and use only the real part of the output.

\item[$abs(QR_X(x))$] Apply quantum rotation on the X-axis and use the absolute values.

\item[$real(QR_Z(x))$] Apply quantum rotation on the Z-axis and use only the real part of the output.

\item[$real(QR_{XYY}(x))$] Apply quantum rotation on all three axes (X, Y, and Z) and use only the real part.

\item[$abs(QR_{XYZ}(x))$] Apply quantum rotation on all three axes (X, Y, and Z) and use the absolute values.

\item[$QR_{XYZ}(x)$] Apply quantum rotation on all three axes (X, Y, and Z) and use both real and imaginary parts.

\item[$real(QR_Z(GN(x)))$] First apply Gaussian noise, then apply quantum rotation on the Z-axis and use only the real part.

\item[$real(QR_Z(F(PR(x))))$] First apply perfect rotation and flipping, then apply quantum rotation on the Z-axis and use only the real part.

\item[$real(QR_Z(C(F(CR(x)))))$] First apply classical rotation, flipping, and cropping, then apply quantum rotation on the Z-axis and use only the real part.

\item[$abs(QR_{XYZ}(C(F(CR(x)))))$] First apply classical rotation, flipping, and cropping, then apply quantum rotation on all three axes (X, Y, and Z) and use the absolute values.

\end{description}

\begin{figure}[ht]
    \centering
    \includegraphics[width=0.8\textwidth]{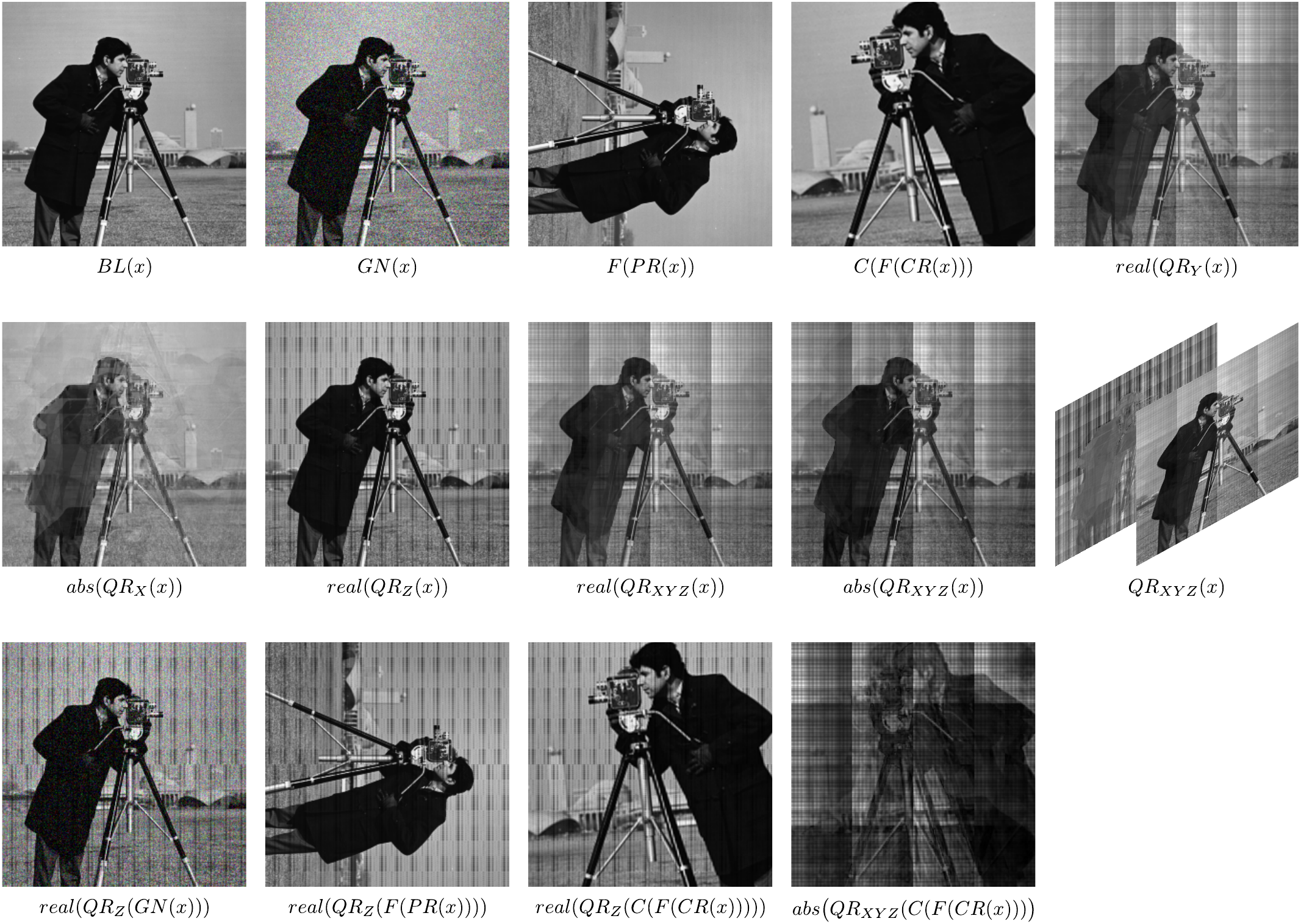}
    \caption{Visualizations of the previously mentioned augmentation methods.}
    \label{fig:augmentations}
\end{figure}

In Fig.~\ref{fig:augmentations}, we use the \texttt{cameraman} image \cite{cameraman} to illustrate the effects of each augmentation method. The classical augmentation methods exhibit expected behaviour; for instance, Gaussian noise introduces a characteristic grainy texture. The outcomes of the other classical techniques are consistent with their intended effects. Turning to the quantum-inspired augmentations, we note that taking the $real(QR_y)$ is equivalent to $QR_y$ as the complex component is zero in this case. For $QR_{XYZ}$ where the data is complex we depict the real and imaginary parts side by side. Furthermore, we avoid $abs(QR_Z)$ as this operation would simply be the identity and $abs(QR_Y)$ would be very similar to $abs(QR_X)$. Notably, local black distortions emerge in augmentations involving $QR_Z$ and block-like structure appears in those involving $QR_Y$ and $QR_{XYZ}$. These visual patterns, along with the reasoning behind our specific choices of methods that combine quantum and classical operations, will be examined in detail in the \hyperref[sec:results]{"Results"} section.

\section*{Methods}\label{sec:m}

\subsection*{Model Architecture}\label{sec:model}
In this work, we utilize the ResNet34 architecture without pre-training, consisting of 34 layers. The model is organized into stages, each featuring residual blocks of increasing complexity. Details and the visualization of the architecture of the model are presented in the work by Kaiming He et al. \cite{he2016deep}. Despite being a relatively older architecture, ResNet34 remains pertinent for various computer vision tasks, including image classification, object detection \cite{girshick2015fast}, and semantic segmentation \cite{chen2018encoder}. While newer models incorporating Transformer architectures have emerged, the simplicity and effectiveness of ResNet make it a still-relevant choice for certain applications.

\subsection*{Normalization} \label{sec:norm}
In line with standard Image Processing practices, we perform a Z-Score normalization on all input images. Before this step, we apply Min-Max Scaling for all quantum augmentations. This preprocessing is necessary due to the impact of quantum augmentation, which transforms the values of each channel in the original image into a range of $]-\infty, \infty[$. Consequently, we apply the subsequent Min-Max normalization procedure. Let $I$ represent an image, and $I^{(j)}$ denote the $j$-th channel of $I$. We use $min(I^{(j)})$ to indicate the minimum value within this channel. Similarly, we define $max(I^{(j)})$. It is worth noting that $min(I^{(j)}) \neq max(I^{(j)})$ invariably holds true. Furthermore, we denote the original image as $I_{org}$ and the quantum-augmented image as $I_{quant}$. With this context, we apply the following 0-1 normalization to map the data to a $[0,1]$ range:
\begin{equation}
    I^{(j)}_{0-1} \coloneqq \frac{I^{(j)}_{quant} - min(I^{(j)}_{quant})}{max(I^{(j)}_{quant})- min(I^{(j)}_{quant})}
\end{equation}
Nevertheless, the original image does not always contain the brightest and darkest possible pixels. Therefore, we rescale this $0-1$ normalized image to the range $[min(I^{(j)}{org}), max(I^{(j)}{org})]$ as follows:
\begin{equation}
    I^{(j)}_{norm} \coloneqq I^{(j)}_{0-1} \cdot \left( max(I^{(j)}_{org})- min(I^{(j)}_{org}) \right) + min(I^{(j)}_{org})
    \label{eq:normalization}
\end{equation}
In case of $QR_{XYZ}(x)$ we have a real and an imaginary part. Therefore we apply the aforementioned Min-Max normalization on both the real and imaginary parts independently.

We also want to mention, that the scaling factor of the normalization will only change singular values of an image by the same factor; however, a constant shift may change the singular values slightly but has no effect on the structure of the image.

\subsection*{Implementation}

To evaluate the effect of our augmentation methods, we design a standard training pipeline based on the ImageNet dataset. First, in the dataloader all figures from the ImageNet dataset are loaded one by one and preprocessed. We then apply the specified augmentation method to each image. For augmentations with multiple augmentation algorithms such as $\mathrm{real}(\mathrm{QR}_Z(\mathrm{GN}(x)))$, the transformations are applied in nested order: First, Gaussian noise ($GN(\cdot)$) is added to the input image $x$, followed by a quantum rotation around the Z-axis ($QR_Z(\cdot)$), and finally, the real part of the resulting image representation is extracted. Afterwards, the augmented data are normalized using min-max scaling, as defined in Eq.~(\ref{eq:normalization}). As usually, those preprocessed images are grouped into batches, where the batch size is dependent on the specific augmentation method and is later optimized through a hyperparameter search. The resulting batches are then used to train ResNet-34. Due to the deprecation of testing interfaces on the official ImageNet website, we evaluate the model performance on the validation dataset. Since we also include the baseline, which uses no augmentations, we are still able compare the different augmentation methods. A genetic algorithm was used  to find reasonable values for hyperparameters as  proposed in \cite{tschope2025novel} with the hyperparameters being batch size, learning rate, $\beta$-values for the Adam optimizer, weight decay, and learning rate decay schedule.

\phantomsection
\section*{Results}\label{sec:results}
\subsection*{Loss and Evaluation Metrics}
\begin{figure}[ht]
    \centering
    \includegraphics[width=0.9\textwidth]{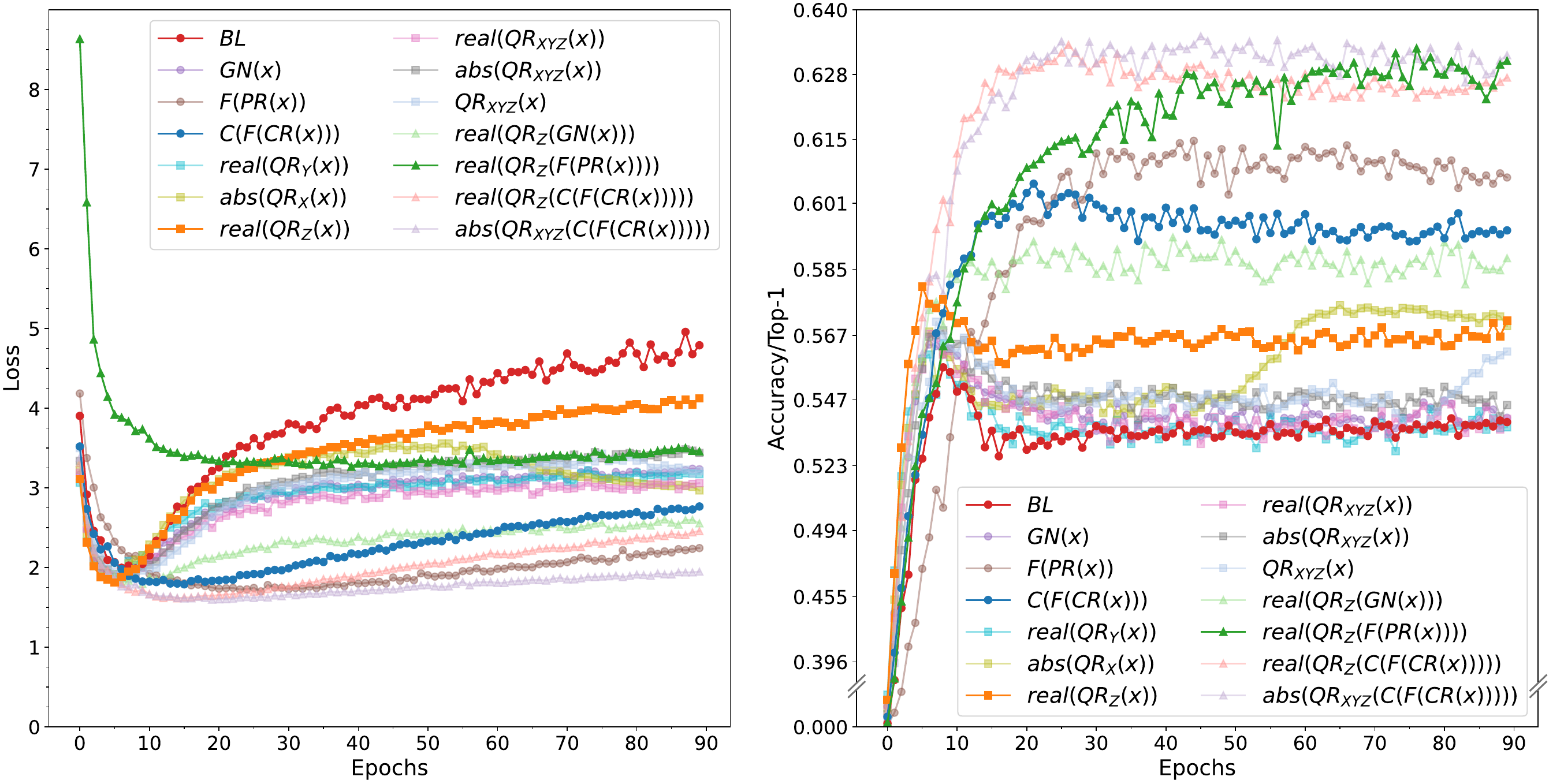}
    \caption{This figure visualizes the losses (left) and accuracies respectively Top-1 scores (right) of all augmentation methods.}
    \label{fig:results}
\end{figure}

In Fig. \ref{fig:results}, we plot the loss (left) and Top-1 accuracy (right) as functions of training epochs for all augmentation methods presented in section \hyperref[sec:am]{"Augmentation methods"}. Upon inspecting the plot, we observe that the most effective classical augmentation appears to be \texttt{$F(PR(x))$}. This method achieves a relatively high accuracy with a rapid upward trend, accompanied by a low loss that decreases quickly before plateauing, indicators of efficient and stable learning. The next best performance among classical methods seems to come from \texttt{$C(F(CR(x)))$}. However, in this case, the accuracy is lower, and the loss, although it initially dips, continues to increase thereafter, suggesting less stable training dynamics.

Concerning the quantum augmentation methods, the method \texttt{$real(QR_Z(x))$} displays the best behaviour, even though its loss curve continues to increase after the initial drop. Good performance is also observed with \texttt{$real(QR_{XYZ}(x))$} and \texttt{$abs(QR_{XYZ}(x))$}. Based on these observations, we proceed to combine the best-performing classical and quantum-inspired augmentation methods. Among the combined methods, \texttt{$real(QR_Z(F(PR(x))))$} demonstrates the most promising behaviour. Specifically, it exhibits a rapidly decreasing and smooth loss curve, alongside a steadily increasing accuracy. Very high accuracy is also achieved by \texttt{$real(QR_Z(C(F(CR(x)))))$} and \texttt{$abs(QR_Z(C(F(CR(x)))))$}, although their loss curves are less stable.

In Table~\ref{tab:results}, we present the results of our model evaluated on the validation dataset. For each augmentation method, we report the \texttt{$F_1$}-Score, along with the Top-1 and Top-5 accuracy metrics. The trends observed in the training loss and accuracy plots are validated by these results. Among the classical methods, \texttt{$F(PR(x))$} consistently achieves the highest scores across all metrics. Similarly, for the quantum-inspired methods, \texttt{$real(QR_z(x))$} outperforms the others. Overall, the combined methods demonstrate superior performance compared to both classical and quantum-inspired augmentations. Notably, \texttt{$real(QR_Z(F(PR(x))))$} achieves the best overall results. Remarkably, its \texttt{$F_1$}-Score is nearly double that of the baseline \texttt{$BL$}, underscoring its effectiveness. Both \texttt{$real(QR_Z(C(F(CR(x)))))$} and \texttt{$abs(QR_Z(C(F(CR(x)))))$} also perform very well. The former even slightly surpasses \texttt{$real(QR_Z(F(PR(x))))$} in terms of Top-1 accuracy, although the difference lies within the margin of error.

Since the performance of $real(QR_Z)$ is the best, we want to understand in detail the effect of $real(QR_Z)$. Given that $R_Z(\theta)$ is diagonal with eigenvalues $\mathrm{e}^{i\theta}$ for eigenstate $\ket{0}$ and $\mathrm{e}^{-i\theta}$ for eigenstate $\ket{1}$ the application of $QR_Z$ from Eq.~(\ref{eq:qrot}) can be rewritten
\begin{align}
\begin{split}
    real(QR_{Z}(\boldsymbol{\theta})) &= real(\bigotimes_{i=0}^{n-1} R_Z^{(i)}(\theta_{Z,i}) \ket{\mathrm{Image}}_{f})\\ &= real\left( \mathrm{e}^{i(\theta_{Z,0} + \theta_{Z,1} +\dots + \theta_{Z,n-1} )},\  \mathrm{e}^{i(-\theta_{Z,0} + \theta_{Z,1} +\dots + \theta_{Z,n-1} )}, \dots, \mathrm{e}^{i(-\theta_{Z,0} - \theta_{Z,1} -\dots - \theta_{Z,n-1}})  \right)^T  \odot \ket{\mathrm{Image}}_{f},  \\
    &= \left( \mathrm{cos}{(\theta_{Z,0} + \theta_{Z,1} +\dots + \theta_{Z,n-1} )},\   \dots, \mathrm{cos}{(-\theta_{Z,0} - \theta_{Z,1} -\dots - \theta_{Z,n-1}})  \right)^T  \odot \ket{\mathrm{Image}}_{f},
\end{split}
\end{align}
where $\odot$ is the Hadamard product (element-wise multiplication) of the two vectors. Since $QR_Z$ is diagonal we can take the real() value separately. Where $real()$ has no effect on the initial $\ket{\mathrm{Image}}_{f}$ it reduces the exponential in $QR_Z$ into a cosine function. The sign of the $\theta_{Z,i}$ are then positional depending on the entry in the qubit register being either $\ket{0}$ or $\ket{1}$. For example, let $n=5$, thus a 5-qubit register state (little endian) of $\ket{01011}=\ket{0}_4\ket{1}_3\ket{0}_2\ket{1}_1\ket{1}_0$, would correspond that $\theta_{z,0}, \theta_{z,1}, \theta_{z,3}$ get a negative sign and the others, where $\ket{0}$ a positive sign. Since we draw the $\theta_{z,i}$ from a random uniform distribution $[ 0, \Theta] $ with mean $\mu=\Theta/2$ and variance of $\sigma^2 = 1/12 \Theta^2$, the sum of $n$ random uniform distributed variables is then Irwin-Hall distributed \cite{hall1927distribution} with mean $\mu=n\Theta/2$ and $\sigma^2 = 1/12 n\Theta$, when all $\theta_{Z_i}$ are positive. Since we have register position depended sign changes of $\theta's$ only the mean will change according to $\mu=(n-2m)\Theta/2$, where $m$ is the amount of $\ket{1}$ in the register state of $n$-qubits. The Irwin-Hall distribution becomes Gaussian distributed with same mean and variance when $n$ becomes large. For our cases it is already a good approximation when $n>10$. This means the application of $real(QR_Z(x))$ corresponds to the application of a register position dependent factor where the corresponding factor corresponds to a $cos(\zeta_k)$, where $\zeta_k$ is drawn from a Gaussian distribution with variance $\sigma^2=1/12 n\Theta$ and $\mu_k=(n-(\#(k)))\Theta/2$, where $\#$ counts the amount 1's in the bit string representation of the integer number $k$. This means the $real(QR_Z)$ augmentation is a positional augmentation, which repeats itself, leading to artifacts in the augmented image which in principle can obscure smaller details in the background, see Fig.~\ref{fig:augmentations}.

Similarly to $QR_Z$ we may address $QR_X$ and $QR_Y$ as for example $R_X = H R_Z H $ and $R_Y = S^\dagger H R_Z HS$. Here $H$ is the Hadamard gate and $S$ the $\pi/2$ phase-gate. Since $QR_X, QR_Y$ are not diagonal, we cannot apply $abs()$ and $real()$ separately. Furthermore, the application of $H$ and $HS$ to the state before and after the application of $R_Z$ shuffles the amplitude among feasible states similarly to the QAOA algorithms to explore the solution space \cite{moll2018quantum, zhou2020quantum}. The feasible states are here qubit dependent, instead of register dependent. We thus see repeating block structures in the augmented pictures. In the example, presented in Fig.~\ref{fig:augmentations} of $n=16$ blocks with repeating embedded sub-blocks of $n=16$, with overall blocks of $16\times16=256$, given the image size of $256\times 256$.

\begin{table}[ht]
\centering
\begin{tabular}{|l|r|r|r|}
\hline
\textbf{Method} & \textbf{F$_1$-Score} & \textbf{Top-1} & \textbf{Top-5} \\
\hline
$BL$ & 6.41\% & 55.75\% & 80.07\% \\ 
$GN(x)$ &  7.09\% & 56.43\% & 80.56\% \\ 
$F(PR(x))$ & 7.93\% & 61.47\% & 84.12\% \\ 
$C(F(CR(x)))$ & 8.20\% & 60.54\% & 82.79\% \\ \hline
$real(QR_{Y}(x))$ & 6.91\% & 56.47\% & 80.25\% \\ 
$abs(QR_{X}(x))$ & 6.74\% & 57.88\% & 81.06\% \\ 
$real(QR_{Z}(x))$ & 8.70\% & 58.05\% & 81.68\% \\ 
$real(QR_{XYZ}(x))$ & 7.16\% & 57.88\% & 80.97\% \\ 
$abs(QR_{XYZ}(x))$ & 7.07\% & 57.14\% & 81.08\% \\ 
$QR_{XYZ}(x)$ & 6.92\% & 57.09\% & 81.12\% \\  \hline
$real(QR_{Z}(GN(x)))$ &  8.53\% & 59.31\% & 82.32\% \\ 
$real(QR_{Z}(F(PR(x))))$ & \textbf{12.51\%} & 63.40\% & \textbf{85.35\%} \\ 
$real(QR_{Z}(C(F(CR(x)))))$ & 9.81\% & 63.36\% & 84.88\% \\ 
$abs(QR_{XYZ}(C(F(CR(x)))))$ & 10.03\% & \textbf{63.52\%} & 85.13\% \\ 
\hline
\end{tabular}
\caption{Results for the classical, quantum and combined augmentations}
\label{tab:results}
\end{table}

\subsection*{Singular Value Spectrum Analysis}
\begin{figure}[ht]
    \centering
    \includegraphics[width=0.9\linewidth]{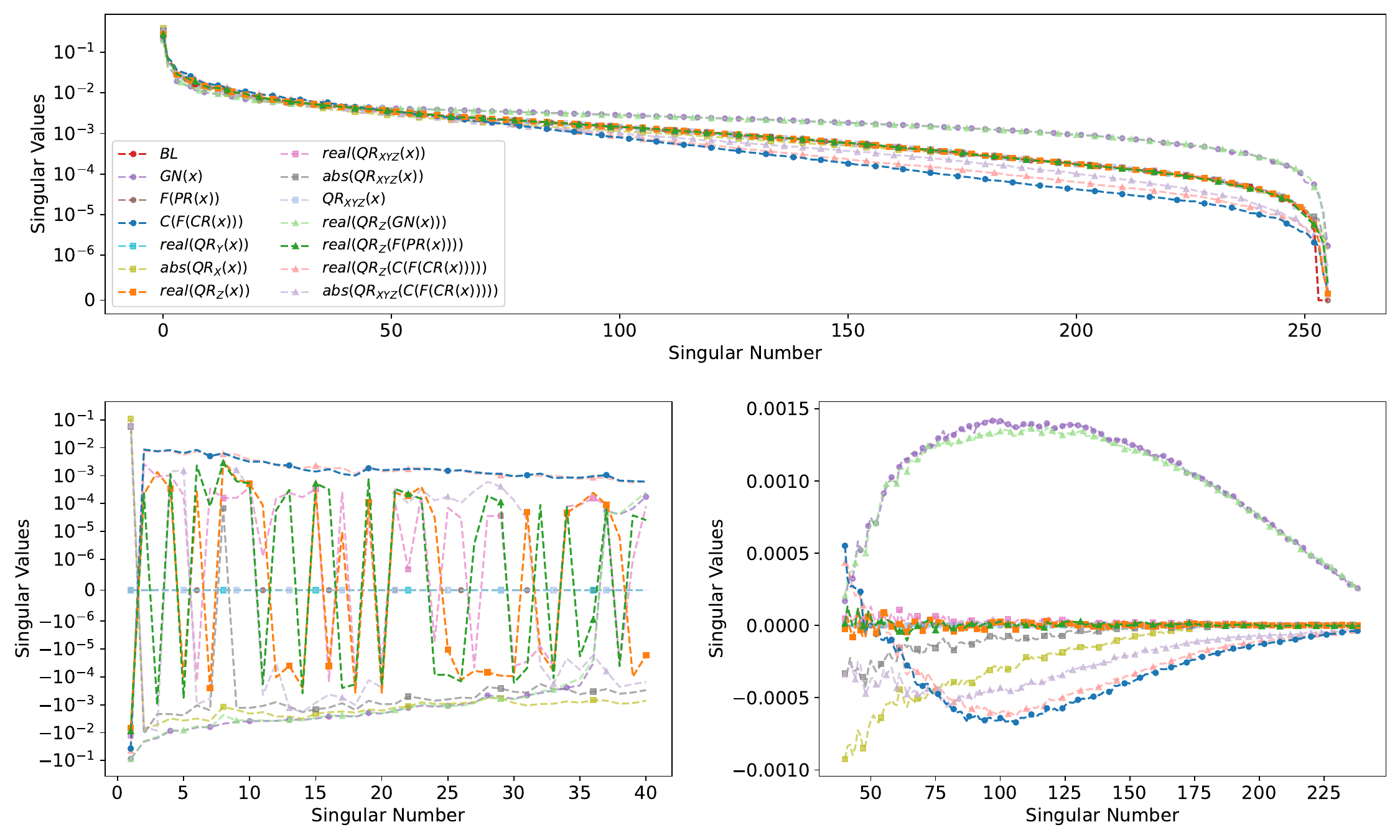}
    \caption{Results of the singular value decomposition for all previously mentioned augmentation methods, including the baseline, applied to the camera man image, already shown in Figure \ref{fig:augmentations}, without the normalization and the unit8 transformation. The figure in the top shows the full range of singular numbers, while the figures in the bottom provide zoomed-in views of specific ranges to highlight distinctions between overlapping curves.}
    \label{fig:Singular_Value_Visualization_Camera_Man}
\end{figure}

\begin{figure}[ht]
    \centering
    \includegraphics[width=0.9\linewidth]{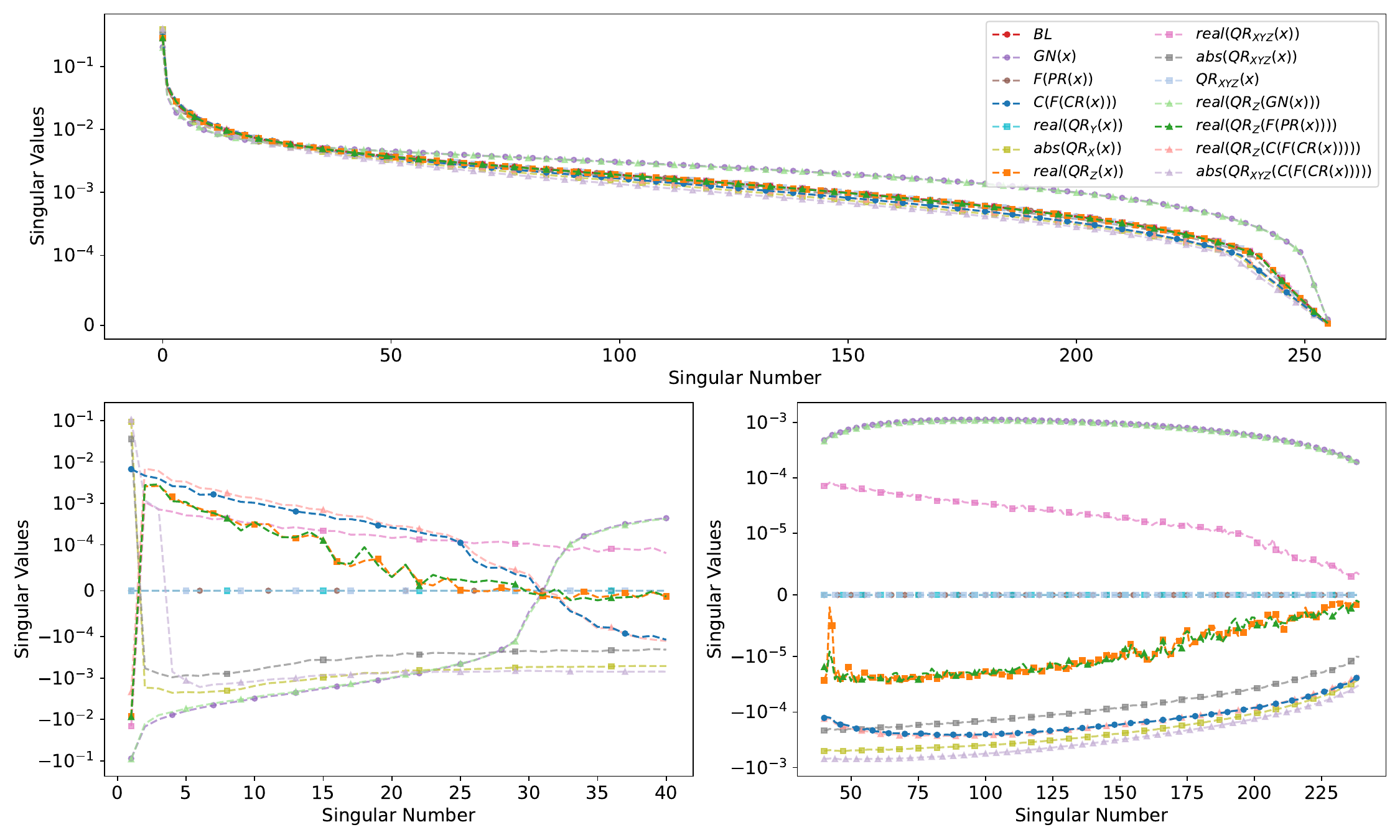}
    \caption{Results of the singular value decomposition averaged over 100 images from the ImageNet dataset using all
augmentation methods, without the normalization and the unit8 transformation. The sub-figure in the top shows the full range of averaged singular values, while the sub-figures in the bottom provide detailed views.}
    \label{fig:Singular_Value_Visualization_100_Images_without_norm_and_without_uint8}
\end{figure}

Next, we compute the singular value decomposition (SVD) and plot the resulting singular value spectra for all augmentation methods. The top panel of Fig.~\ref{fig:Singular_Value_Visualization_Camera_Man} presents the spectra for the \texttt{cameraman} image, while Fig.~\ref{fig:Singular_Value_Visualization_100_Images_without_norm_and_without_uint8} shows the average spectra computed over 100 images from the ImageNet dataset without normalization and without unit8 transformation. The bottom-left and bottom-right panels of each figure display zoomed-in views of selected regions from the corresponding top panel, allowing for closer inspection of differences among the augmentation methods. In these bottom panels, however, we plot the difference between each augmentation method and the baseline, rather than the raw spectra. This highlights how each transformation alters the singular value distribution relative to the unaugmented input.

The singular value spectra provide insight into how each augmentation method affects the underlying structure of the data by changing the importance on either key-features of the image (high-valued singular values) or sub-features which otherwise would be hidden within the image. Specifically, a sharp decay in the spectrum indicates that most of the variance is concentrated in the first few singular values, suggesting that the image structure is dominated by a few principal components. In contrast, a slow decay implies that the information is more evenly distributed across many components, which may indicate, for example, increased complexity, or a loss of dominant structure.

We note that \texttt{$F(PR(x)$}, in both figures, sits on top of the baseline as expected, indicating that the singular values remain unchanged and the same is the case for the quantum-inspired augmentations $real(QR_Y(x))$ and $QR_{XYZ}$. However, in the quantum case there will be a small discrepancy which is due to datatype changes from float64 (complex128) to uint8 and the normalization, which includes an addition to push negative numbers back to positive. In the \hyperref[sec:sup]{"Supplementary B"} we show the corresponding images with uint8 and normalization application.

Both \texttt{$real(QR_Z(x))$} and \texttt{$real(QR_Z(F(PR(x)))$} exhibit similar behaviour. In Fig.~\ref{fig:Singular_Value_Visualization_Camera_Man}, the initial singular values are lower than those of the baseline, followed by pronounced oscillations that gradually diminish as the curve converges toward the baseline. We note that when the average over 100 images is considered, as shown in Fig.~\ref{fig:Singular_Value_Visualization_100_Images_without_norm_and_without_uint8}, the oscillations in the singular value spectra are significantly smoothed out. The fact that the first singular values are lower than those of the baseline indicates that less variance is captured by the leading principal components. This may correspond to a reduction in dominant image structure, which appears to be characteristic of the way the $real(QR_Z(x))$ transformation affects the data. The black vertical local and repeating distortions (artifacts) observed in Fig.~\ref{fig:augmentations} are a visual manifestation of this structural alteration.

The augmentation methods that incorporate Gaussian noise (\texttt{$GN(x)$} and \texttt{$real(QR_Z(GN(x)))$}) exhibit a consistent pattern across both the singular value spectrum of a single image (Fig.~\ref{fig:Singular_Value_Visualization_Camera_Man}) and the average over 100 images (Fig.~\ref{fig:Singular_Value_Visualization_100_Images_without_norm_and_without_uint8}). This consistency is expected, given the stochastic yet statistically uniform nature of Gaussian noise. In both cases, the leading singular values fall below the baseline, indicating a loss of dominant structure. This is visually manifested as a grainy appearance (see Fig.~\ref{fig:augmentations}), which is a characteristic effect of Gaussian noise. Afterwards, the singular values increase smoothly and rapidly, rising above the baseline and forming a broad plateau. Toward the higher-order components, we observe a tendency for the spectrum to return toward the baseline.

In Fig.~\ref{fig:Singular_Value_Visualization_100_Images_without_norm_and_without_uint8}, we observe that the classical augmentation \texttt{$C(F(CR(x)))$} exhibits behaviour similar to the combined method \texttt{$real(QR_Z(C(F(CR(x)))))$}. In both cases, the initial singular values are higher than those of the baseline, indicating an enhancement of the first principal components. Following this, the spectral curve gradually diminishes, drops below the baseline, and eventually begins to rise again. In Fig.~\ref{fig:Singular_Value_Visualization_Camera_Man}, we observe more or less a similar behaviour, with two notable exceptions: the first singular values start below the baseline, and the spectral curve appears less smooth. These deviations are artifacts of using a single image and a single crop.

The remaining augmentation methods exhibit a pattern in which the initial singular values are significantly higher than the baseline, followed by a sudden drop below it. This is typically followed by an extended plateau, with a slight tendency to return toward the baseline at the tail end of the spectrum.

\section*{Discussion}\label{sec:discuss}
In general, our findings suggest that quantum-inspired augmentations, particularly $real(QR_Z(x))$ and the best performing augmentation $real(QR_Z(F(PR(x))))$, lead to changes in the singular value spectrum that closely resemble those produced by classical cropping and rotation augmentations $C(F(CR(x)))$, but without actual loss of pixel values. Furthermore, $real(QR_Z(x))$ provides a positional augmentation which visually manifests as a repeating structural alteration of the image. This indicates that the transformations may act by effectively obscuring portions of the input image, analogous to dropout-like effects in classical models. Furthermore, $real(QR_Z)$ does not yield a significant overhead but can effectively be computed in $\mathcal{O}(N)$.

When we compare each augmentation method with the baseline
$BL(x)$, we obtain the following improvements. For the classical augmentation $C(F(CR(x)))$, the F$_1$-score increases by $1.8\%$, Top-1 by $4.8\%$, and Top-5 by $2.7\%$. For the quantum-inspired augmentation $\text{real}(QR_Z(x))$, the F$_1$-score and Top-1 both increase by $2.3\%$, while Top-5 improves by $1.6\%$. The best-performing method, $\text{real}(QR_Z(F(PR(x))))$, achieves an increase of $6.1\%$ in the F$_1$-score, $7.6\%$ in Top-1, and $5.2\%$ in Top-5. When comparing the best-performing to the best classical augmentation method, this gain corresponds to an improvement of 3\% in Top-1, a 2.5\% in Top-5 and an increase of F$_1$ score from 8\% to 12\%. Interestingly, many of the quantum-inspired and combined augmentations appear to emphasize the leading singular values and introduce spectral profiles that mimic Gaussian noise in the mid spectrum while diverging at the tail. These special characteristics may explain the observed improvements in model generalization and robustness.

Importantly, our analysis reveals that quantum rotations which preserve the singular value spectrum, such as $real(QR_Y(x))$ or more complex unitary operations like $QR_{XYZ}(x)$, do not yield significant performance gains when compared to the baseline. This highlights that the key benefit arises not from the quantum operation itself, but rather from its non-unitary classical projection, such as the application of of $abs()$ or $real()$, which introduce non-trivial distortions in the data space. These effects are amplified when combined with conventional classical augmentations, pointing to a promising method for designing future augmentation pipelines that blend interpretable quantum-inspired components with robust classical methods.

In conclusion, our work demonstrates that combining classical with quantum-inspired augmentations can provide effective and interpretable tools for enhancing data diversity in classical ML tasks—offering both theoretical insights and practical benefits.

\FloatBarrier
\section*{Funding}
This work was funded by the Quanten-Initiative Rheinland-Pfalz (QUIP), Research Initiative Quantum Computing for AI (QC-AI), and F\"orderkennzeichen Open6GHub 16KISK003K.

\section*{Data availability}
Data and Code will be made available on reasonable request. Correspondence and requests for materials should be addressed to M.K-E.



\section*{Acknowledgements}

We gratefully acknowledge financial support from the Quantum Initiative Rhineland-Palatinate QUIP and the Research Initiative Quantum Computing for AI (QC-AI). This work has been supported by F\"orderkennzeichen Open6GHub 16KISK003K.

\section*{Supplementary Material}\label{sec:sup}

\section*{Supplementary A: Differential Privacy}
We began by investigating whether the quantum-inspired augmentations alone could offer privacy benefits akin to Differential Privacy (DP), given their ability to obscure visual features through structured randomness. At first glance, the visual distortion introduced by these transformations suggested a potential for privacy preservation as these augmentations are largely information-preserving—particularly in terms of the singular value spectrum, which remains mostly intact.

In 2013, Dwork et al. \cite{dwork2014algorithmic, dwork2006calibrating} introduced the concept of DP. This concept is designed to ensure that the output of an algorithm remains statistically indistinguishable whether or not any individual data point is included in the input dataset. It provides quantifiable privacy guarantees by bounding the risk of inferring sensitive information about any single individual from the released results. This is typically achieved by including random noise into either the data or the output of computations, controlled by a privacy parameter $\varepsilon$ (epsilon), where smaller values imply stronger privacy. DP is also a foundational concept in privacy-preserving machine learning, enabling the use of sensitive data-such as personal images or medical data, while minimizing the risk of re-identification or data leakage. 
To test our methods in the context of differential privacy, we used the KoIn (Korean Influencer) dataset \cite{seo2023new}. This dataset is a facial image dataset specifically for benchmarking face recognition models under real-world conditions. It contains real-world photos from more than $100,000$ Korean influencers. There are various cases to train and test on this dataset. We use the dataset with $57$ classes, where each class contains approximately $1000$ images. We train and test a ResNet34 \cite{he2016deep} on the predefined training and test set.

\begin{figure}[ht]
    \centering
    \includegraphics[width=0.6\textwidth]{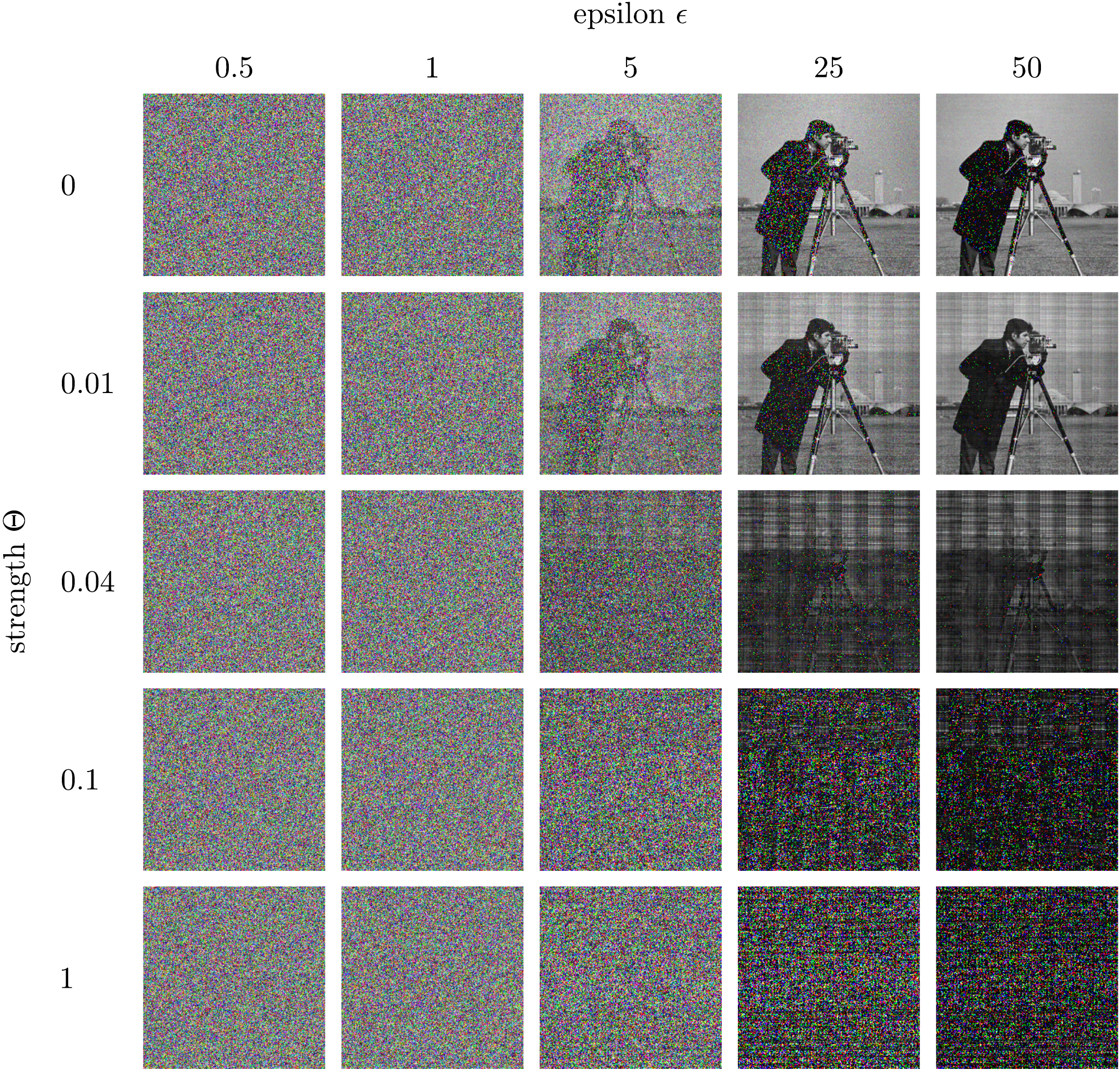}
    \caption{Overview of differential privacy with $real(QR_{XYZ}(x))$ augmentation, using various epsilon and strength values.}
    \label{fig:DP_with_and_without_QA}
\end{figure}
DP is often defined using the concept of neighbouring datasets. Let $D$ and $D'$ be two datasets that differ by at most one data point, and $A$ be a randomized algorithm (e.g., a machine learning model or the quantum-inspired augmentations in this paper). The algorithm $A$ is said to satisfy $\varepsilon$-DP if, for all possible neighbouring datasets $D$ and $D'$, and for all subsets of possible outcomes $S$,
$$
\Pr[A(D) \in S] \leq e^\varepsilon \cdot \Pr[A(D') \in S].
$$

Here, $\varepsilon$, the privacy budget, serves as a measure of how much privacy is preserved in the process of using the algorithm $A$. A privacy budget can be thought of as a quantifier of the amount of information leakage allowed by the algorithm $A$. With a small value of $\varepsilon$, usually between $0$ and $1$, it can be guaranteed that the results of $A$ won't reveal too much about any individual's data, even if they were part of the dataset \cite{dwork2006calibrating,dwork2006differential}.

($\varepsilon$, $\delta$)-DP is a relaxation of $\varepsilon$-DP that introduces an additional parameter $\delta$, which represents the maximum allowable probability of a noticeable deviation from perfect privacy \cite{dwork2006our}. This formulation is more flexible and allows for a small probability ($\delta$) of failing to meet the $\varepsilon$-privacy requirement. The guarantee is that the algorithm's output does not reveal too much information about any individual, except with a small probability $\delta$. Formally:
$$ \text{Pr}[A(D) \in S] \leq e^\varepsilon \times \text{Pr}[A(D') \in S] + \delta $$

Besides $\varepsilon$-DP and ($\varepsilon$, $\delta$)-DP, there are other formulations and extensions of DP that cater to specific privacy needs or constraints. These include Rényi Differential Privacy \cite{mironov2017renyi} and Gaussian Differential Privacy \cite{dong2022gaussian}. Each formulation has its strengths and is applicable in specific scenarios depending on the sensitivity of the data and the desired level of privacy protection.

DP can be applied to generative models for images to protect individuals' privacy while maintaining utility and visual quality. In this context, the traditional concepts of adjacency and query sensitivity are replaced by a notion of distance between secrets, where a secret represents the numeric representation of an individual. This distance metric is crucial for ensuring that similar secrets are indistinguishable while maintaining distinguishability for very different secrets.

For image representations within generative models, distance can be calculated between images by mapping them to fixed-length vectors. The application of DP to generative neural networks involves representing an image as a vector $X$ in $\mathbb{R}^n$ and applying a randomization mechanism $K: \mathbb{R}^n \rightarrow \mathbb{R}^n$ to produce an obfuscated instance.

The privacy guarantee is adapted to this vector representation by introducing a distance function $d: \mathbb{R}^n \times \mathbb{R}^n \rightarrow \mathbb{R}$. The DP generalization is then expressed as:
$$ \text{Pr}[K(X_1) = R] \leq e^{\varepsilon \cdot d(X_1, X_2)} \text{Pr}[K(X_2) = R] \quad \forall X_1, X_2, R \in \mathbb{R}^n, $$
where $d(X_1, X_2)$ represents the distance between $X_1$ and $X_2$ according to the chosen metric \cite{ji2022privacy}.

A suitable choice for the distance metric is the $L_1$ distance, normalized to ensure all elements fall within the range $[0, 1]$. This normalization accounts for differences in the ranges of elements in the vectors. The final distance metric sums the normalized distances for each element of the vectors, providing a meaningful privacy guarantee for image representations across different models.

Images are increasingly becoming a vital part of our digital lives, capturing personal information, medical data, and sensitive details. As advancements in image processing and augmentation techniques unlock exciting possibilities, protecting privacy within these applications becomes important. This is where differential DP emerges as a powerful technique, offering rigorous mathematical guarantees to protect privacy while enabling valuable image processing tasks. However, the quantum inspired image augmentation is not differentially private.

\noindent\textbf{Theorem:} The quantum-inspired image augmentation is not differentially private.

\noindent\textbf{Proof:}
We prove that there exist images $\Psi_1,\Psi_2$ and $R$ such that 
$$
\text{Pr}[\mathbf{R}(\Theta)\ket{\Psi_1} = R]=0 \quad \text{and} \quad \text{Pr}[\mathbf{R}(\Theta)\ket{\Psi_2} = R]=1
$$
for any $\Theta$, which implies that $\mathbf{R}(\Theta)$ is not DP. Choose $\Psi_1$ and $\Psi_2$ such that 
$$
\mathbf{R}(\Theta)\ket{\Psi_1} \neq \mathbf{R}(\Theta)\ket{\Psi_2}.
$$
This choice is possible because the rectangular matrix (with singular values on the diagonal) of the singular value decomposition of $\mathbf{R}(\Theta)\ket{\Psi}$ and $\Psi$ are the same. Let $R = \mathbf{R}(\Theta)\ket{\Psi_1}$. Then, we have for any $\Theta$:
\begin{align*}
    \text{Pr}[\mathbf{R}(\Theta)\ket{\Psi_1} = R] &= 1, \\
    \text{Pr}[\mathbf{R}(\Theta)\ket{\Psi_2} = R] &= 0.
\end{align*}
Supposing that $\mathbf{R}(\Theta)$ is $\varepsilon$-DP, we have:
$$
\text{Pr}[\mathbf{R}(\Theta)\ket{\Psi_1} = R] \leq e^{\varepsilon d(\Psi_1,\Psi_2)} \cdot \text{Pr}[\mathbf{R}(\Theta)\ket{\Psi_2} = R],
$$
which implies that: $ 1 \leq 0, $ a contradiction.

Now, assume that $\mathbf{R}(\Theta)$ is $(\varepsilon,\delta)$-DP. Then, we have:    $$\text{Pr}[\mathbf{R}(\Theta)\ket{\Psi_1} = R] \leq e^{\varepsilon d(\Psi_1,\Psi_2)} \cdot \text{Pr}[\mathbf{R}(\Theta)\ket{\Psi_2} = R] + \delta,$$ 
    $$1 \leq 0 + \delta \Rightarrow \delta \geq 1.$$

However, since $\delta \leq 1$ by definition, this means that there is a probability of $1$ of failing to meet privacy. Hence, $\mathbf{R}(\Theta)$ is not DP.

$\mathbf{R}(\Theta)$ fails to meet the formal requirements of DP. To better understand the interaction between these two mechanisms, we examined the effect of applying DP directly to quantum-augmented images. Further examination showed that the $real(QR_Z(x))$ transformation tends to amplify singular values in the mid-spectrum-regions of the image's spectral profile that are typically less susceptible to noise perturbations. In contrast, noise introduced for DP primarily impacts the tails of the spectrum, reducing their influence but leaving the mid-spectrum relatively unaltered. This indicates that while quantum augmentation alters visual features and can simulate encrypted appearance, it does not mask the underlying data in a way that satisfies formal DP definitions.

Table \ref{tab:DP_results} summarizes our results on the KoIn dataset. As a baseline, we first evaluated the performance without any augmentation. Afterwards, we tested various combinations of quantum augmentations followed by DP. These are denoted as $DP_{\varepsilon}(abs(QR_{Z,\Theta}(x)))$, where $\varepsilon$ relates to the privacy parameter of the DP algorithm, and $\Theta$ represents the strength used in the quantum augmentation, while the sensitivity is for all tests 255. Additionally, we considered a setting in which a strong quantum rotation (more specifically $\Theta = 1$) was applied using a fixed random seed. This scenario simulates the potential use of quantum rotations for image encryption, where the resulting noise matrix could serve as a secret key. While Figure \ref{fig:DP_with_and_without_QA} visualizes the noise-level by changing the values $\Theta$ and $\varepsilon$, Table \ref{tab:DP_results} shows that applying a quantum rotation with $\Theta = 1$ leads to results that lie between those of $DP_{1}(x)$ and $DP_{5}(x)$.

\begin{table}[ht]
\centering
\begin{tabular}{|l|r|r|r|}
\hline
\textbf{Method} & \textbf{F$_1$-Score} \\
\hline
BL & 59.99\% \\ 
\hline
$DP_{0.5}(abs(QR_{Z,0.01}(x)))$ & 1.71\% \\ 
$DP_{1}(abs(QR_{Z,0.01}(x)))$ & 3.71\% \\ 
$DP_{5}(abs(QR_{Z,0.01}(x)))$ & 30.45\% \\ 
$DP_{25}(abs(QR_{Z,0.01}(x)))$ & 49.66\% \\ 
$DP_{50}(abs(QR_{Z,0.01}(x)))$ & 55.08\% \\
\hline
$DP_{0.5}(abs(QR_{Z,0.15}(x)))$ & 1.96\% \\ 
$DP_{1}(abs(QR_{Z,0.15}(x)))$ & 1.79\% \\ 
$DP_{5}(abs(QR_{Z,0.15}(x)))$ & 24.03\% \\ 
$DP_{25}(abs(QR_{Z,0.15}(x)))$ & 47.05\% \\ 
$DP_{50}(abs(QR_{Z,0.15}(x)))$ & 50.84\% \\
\hline
$abs(QR_{Z,1}(x))$ & 34.46\% \\
\hline
$DP_{0.5}(x)$ & 7.69\% \\ 
$DP_{1}(x)$ & 13.94\% \\ 
$DP_{5}(x)$ & 41.86\% \\ 
$DP_{25}(x)$ & 51.54\% \\ 
$DP_{50}(x)$ & 57.91\% \\
\hline
\end{tabular}
\caption{Face classification results on the KoIn dataset with different combinations of quantum augmentation and differential privacy.}
\label{tab:DP_results}
\end{table}

\FloatBarrier

\section*{Supplementary B: Additional Results}
\begin{figure}[ht]
    \centering
    \includegraphics[width=0.9\linewidth]{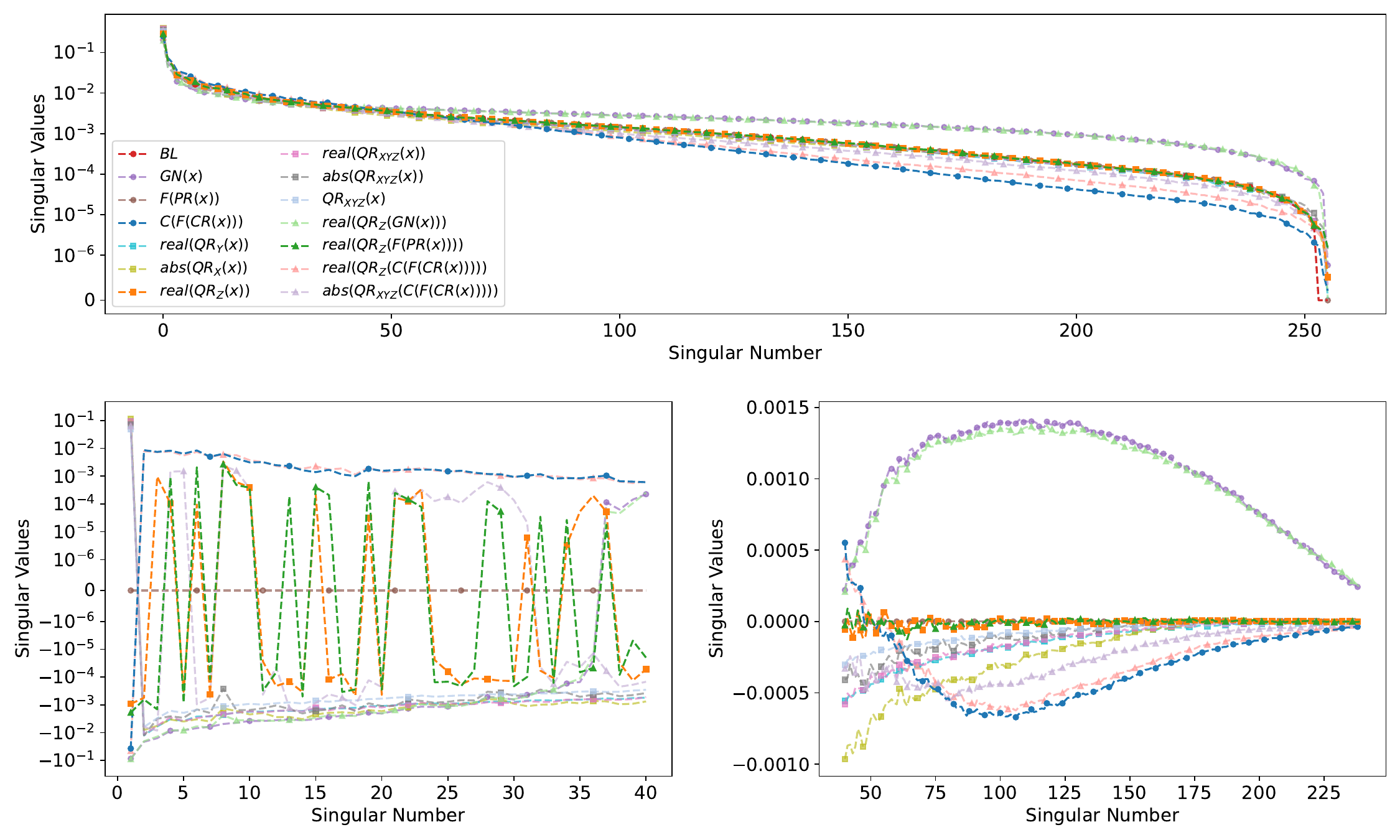}
    \caption{Results of the singular value decomposition for all augmentation methods, including the baseline, applied to the \texttt{cameraman} image. Normalization and the datatype conversion are applied after the augmentation. The figure in the top shows the full range of singular numbers, while the figures in the bottom provide zoomed-in views of specific ranges to highlight distinctions between overlapping curves.}
    \label{fig:SVD_app1}
\end{figure}

\begin{figure}[ht]
    \centering
    \includegraphics[width=0.9\linewidth]{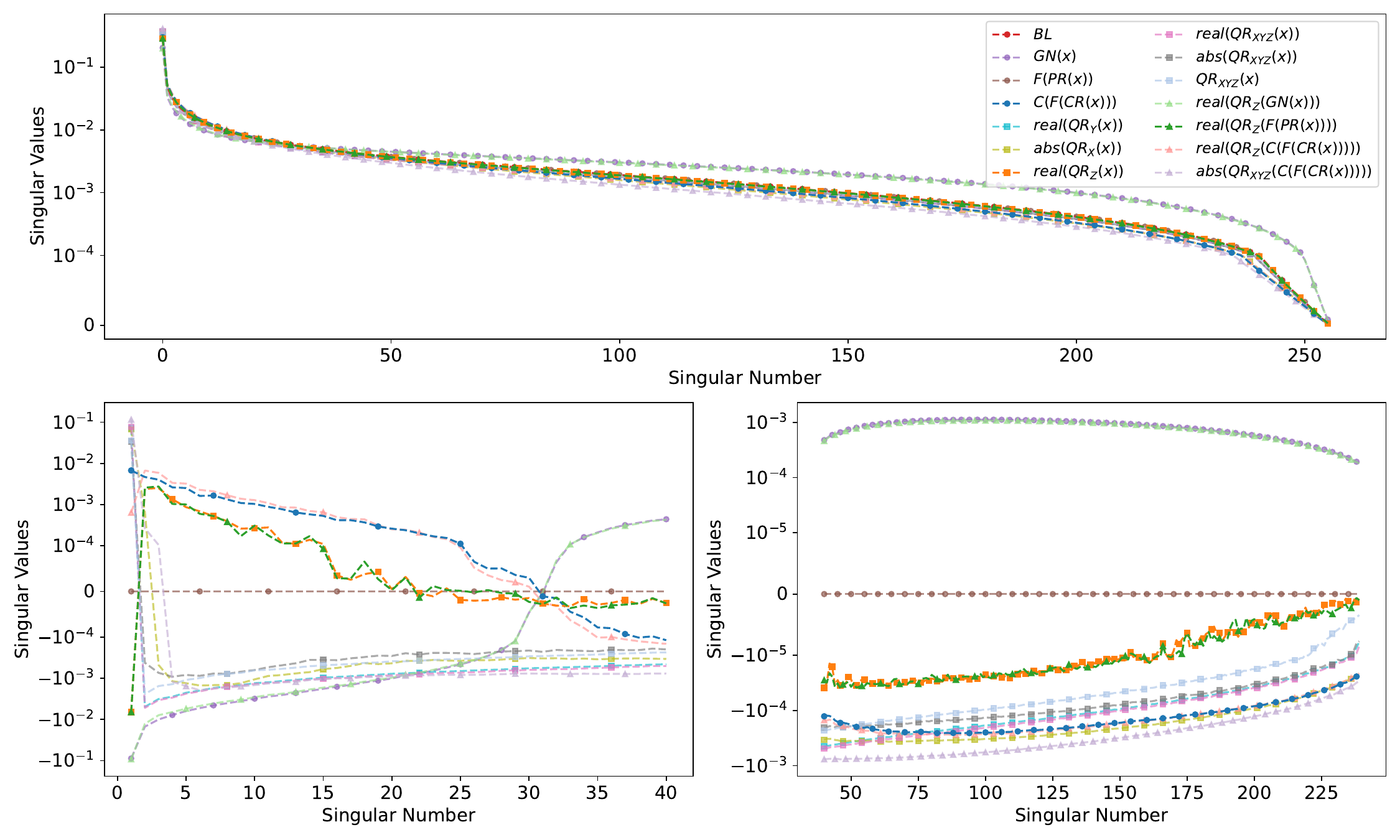}
    \caption{Results of the singular value decomposition averaged over 100 images from the ImageNet dataset using all augmentation methods. Normalization and the datatype conversion are applied after the augmentation.The sub-figure in the top shows the full range of averaged singular values, while the sub-figures in the bottom provide detailed views.}
    \label{fig:SVD_app2}
\end{figure}

In this supplementary we present two figures that correspond to Fig. 5 and Fig. 6 from the main manuscript, with the key difference that both the normalization step and the float-to-unit8 datatype conversion are here taken into account. As a result, in both plots (Fig.~\ref{fig:SVD_app1} and Fig.~\ref{fig:SVD_app2}) we observe that the singular value spectral curve for the quantum augmentations $real(QR_Y(x))$ and $QR_{XYZ}(x)$ deviate from the baseline curve. On the contrary, this does not occur for the \texttt{$F(PR(x)$} spectral curve as neither normalization nor datatype conversion affects the augmentation in this case. 

\section*{Author contributions statement}

M.T. performed all calculations (including the implementation of the classical augmentations), created the figures, and contributed significantly to the writing of the initial manuscript. M.K.-E. developed the quantum augmentation methods, provided key insights, and drafted the first version of the manuscript. M.T., M.K.-E., V.F.R., S.P.S., N.P., and P.L. contributed to the interpretation of the results and offered critical insights throughout the study. S.P.S. provided the theoretical proof and expert input on differential privacy. All authors thoroughly reviewed the manuscript and approved it for publication.

\section*{Additional information}
\noindent \textbf{Declaration of Interest}: The authors declare no competing interests.

\noindent \textbf{AI Tools} like chatGPT4o, DeepL Writer, Grammarly, Writefull(TeXGPT) have been used only to enhance spelling, grammar as well as shortening long sentences. All improved text has been carefully read and made sure to be free of hallucinations.

\end{document}